\documentclass[10pt,twocolumn,letterpaper]{article}

\usepackage{iccv}
\usepackage{times}
\usepackage{epsfig}
\usepackage{graphicx}
\usepackage{amsmath}
\usepackage{amssymb}

\usepackage{subfigure}
\usepackage{booktabs}
\usepackage{dsfont}
\usepackage{graphicx} % more modern
\usepackage{algorithm}
\usepackage{algorithmic}
\usepackage{caption}
\usepackage{mathtools}
\usepackage{stackengine} % eq align
\usepackage{bbm}
\usepackage{siunitx}
\usepackage{amsmath}
\usepackage{pifont}% 
\newcommand{\cmark}{\ding{51}}
\newcommand{\xmark}{\ding{55}}
\usepackage{mathrsfs} % \mathscr
\usepackage{makecell} % change line in table

\usepackage[breaklinks=true,bookmarks=false]{hyperref}

\iccvfinalcopy % *** Uncomment this line for the final submission

\def\iccvPaperID{6899} % *** Enter the ICCV Paper ID here

\ificcvfinal\fi

\begin{document}

%%%%%%%%% TITLE
\title{ Revisiting Image Reconstruction for Semi-supervised Semantic Segmentation}

\author{Yuhao Lin, Haiming Xu, Lingqiao Liu, Jinan Zou, Javen Qinfeng Shi\\
Australian Institute for Machine Learning\\
University of Adelaide\\
{\tt\small \{yuhao.lin01, hai-ming.xu, lingqiao.liu, jinan.zou, javen.shi\}@adelaide.edu.au}
% Haiming Xu\\
% Australian Institute for Machine Learning\\
% University of Adelaide\\
% {\tt\small hai-ming.xu@adelaide.edu.au}
% Lingqiao Liu\\
% Australian Institute for Machine Learning\\
% University of Adelaide\\
% {\tt\small lingqiao.liu01@adelaide.edu.au}
% Jinan Zou\\
% Australian Institute for Machine Learning\\
% University of Adelaide\\
% {\tt\small jinnan.zou@adelaide.edu.au}
% Javen Qinfeng Shi\\
% Australian Institute for Machine Learning\\
% University of Adelaide\\
% {\tt\small javen.shi@adelaide.edu.au}
}

\maketitle
\ificcvfinal\thispagestyle{empty}\fi

%%%%%%%%% ABSTRACT
\begin{abstract}
   Autoencoding, which aims to reconstruct the input images through a bottleneck latent representation, is one of the classic feature representation learning strategies. It has been shown effective as an auxiliary task for semi-supervised learning but has become less popular as more sophisticated methods have been proposed in recent years. In this paper, we revisit the idea of using image reconstruction as the auxiliary task and incorporate it with a modern semi-supervised semantic segmentation framework. Surprisingly, we discover that such an old idea in semi-supervised learning can produce results competitive with state-of-the-art semantic segmentation algorithms. By visualizing the intermediate layer activations of the image reconstruction module, we show that the feature map channel could correlate well with the semantic concept, which explains why joint training with the reconstruction task is helpful for the segmentation task. Motivated by our observation, we further proposed a modification to the image reconstruction task, aiming to further disentangle the object clue from the background patterns. From experiment evaluation on various datasets, we show that using reconstruction as auxiliary loss can lead to consistent improvements in various datasets and methods. The proposed method can further lead to significant improvement in object-centric segmentation tasks.
\end{abstract}

%%%%%%%%% BODY TEXT
\section{Introduction}
Autoencoding aims to reconstruct inputs as outputs with the least possible amount of distortion\cite{baldi2012autoencoders} through an information bottleneck created with low dimension or low-resolution latent variables. Because of its simplicity and effectiveness, it has attracted researchers' attention since it was first introduced in the 1980s\cite{rumelhart1985learning}. Autoencoders again enter the visions of the researchers when the deep stacked-autoencoder architectures\cite{hinton2006reducing} have shown state-of-the-art results as a feature extractor. Nowadays, as one of the most classic representation learning strategies, autoencoder has been widely applied in different applications, such as clustering \cite{guo2017deep}, and classification \cite{luo2017convolutional}.
It has also been discovered \cite{le2018supervised,ghifary2016deep}  that an auto-encoder style reconstruction task could be an excellent auxiliary task for semi-supervised learning. In semi-supervised learning settings, we have access to a large volume of unlabelled data and only a small number of labeled training samples. The reconstruction objective, however, can be applied without any class labels. However, only a small number of labeled training samples and the reconstruction loss can be trained without any class labels. Such a scheme has become less popular as more sophisticated methods \cite{chen2021semi, hu2021semi, liu2021perturbed} have been proposed recently. 

Semi-supervised semantic segmentation is a challenging yet important topic in computer vision, with many real-world applications. It requires utilizing both labeled and unlabeled data to improve segmentation results. As an important application of semi-supervised learning, many semi-supervised learning methods \cite{sohn2020fixmatch, hyun2020class} have been applied and extended to solve the semi-supervised segmentation problem \cite{yun2019cutmix, wang2022semi, hu2021semi}. However separating the fuzzy margin between foreground and background is still challenging.
%-------------------------------------------------------------------------

\begin{figure*}[t!]
    \centering
    \includegraphics[width=\textwidth]{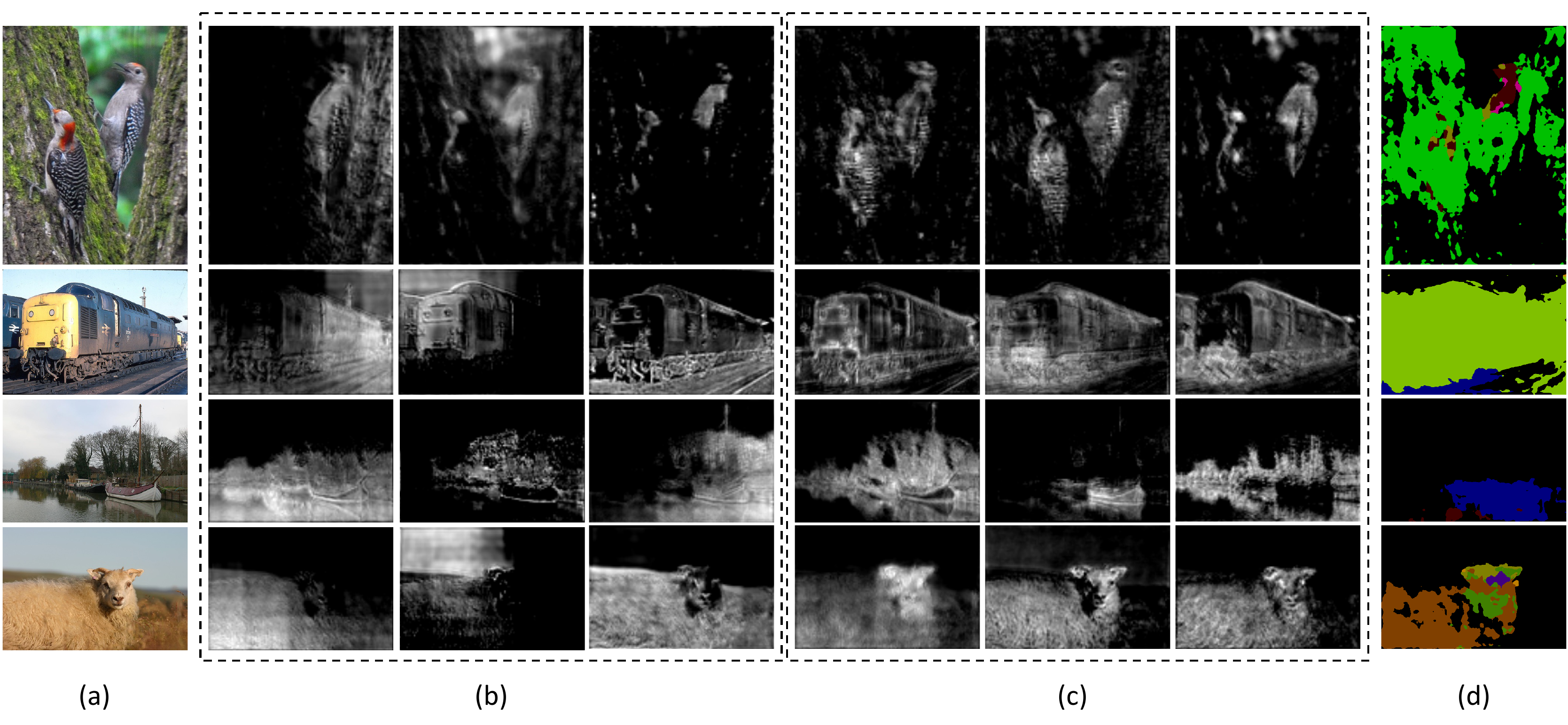}

    \caption{Some visualization results from Pascal VOC 2012 \textbf{validation set} (the reconstruction branch is only trained on the training set). (a) input images, (b) feature maps from the reconstruction-only model, (c) feature maps from the reconstruction-segmentation model, and (d) the segmentation results of the same epoch as c. We observe that the activation areas in some feature maps are focused on the objects (i.e., birds in the first row and goats in the fourth row). 
    } 

    \label{fig:quality_results}
\end{figure*}

This work explores the use of an autoencoder-style reconstruction task to improve semi-supervised segmentation. Perhaps surprisingly, we find that if we incorporate the image reconstruction task with a commonly used semi-supervised segmentation baseline method \cite{lee2013pseudo,yun2019cutmix}, the final performance can be improved, especially when the number of training images is small (see Figure \ref{ab}). This motivates us to understand how reconstruction helps the segmentation task. By visualizing the intermediate activation maps of the reconstruction branch (see Figure \ref{fig:quality_results}), we find that the latent activations of a reconstruction branch have already uncovered the semantics of objects if the reconstruction branch is jointly trained with a semi-supervised segmentation loss. This explains the benefit of the reconstruction task for semi-supervised segmentation, as both tasks are shared with some similarities. From further observation of the latent activations, we noticed that the object and part of its background could often co-occur in one feature map, suggesting potential entanglement of the object and background clue. Thus, we propose a strategy to further disentangle those two clues and expect to align the reconstruction task and segmentation task better. Specifically, we propose to reconstruct foreground-region-only images for the labeled image and perform a similar reconstruction task but apply the loss to partial pixels, as guided by the pseudo-label generated from the segmentation head. Through our experimental evaluation on various datasets, we show that joint reconstruction can be used as a strong semi-supervised segmentation baseline that achieves consistent improvement under different scenarios. 

The main contributions of this paper are highlighted as follows:
\begin{itemize}
    \item We revisit image construction as an auxiliary task for semi-supervised segmentation and show that it can be very effective when working together with existing semi-supervised segmentation methods.
    \item We visualize the intermediate activations of the reconstruction decoder and shed light on why it is beneficial for the segmentation task.
    \item We further propose a method that modifies the reconstruction task and makes the reconstruction tasks more suited for the object-centric segmentation problems.
\end{itemize}

\section{Related Work}

\subsection{Autoencoders for Semi-Supervised Learning}

As one classic unsupervised representation learning approach, Autoencoders (AE) are widely used for unsupervised learning and as a regularization scheme in semi-supervised learning\cite{baldi2012autoencoders,kang2020autoencoder}. Because of the simplicity, they have been attracting researchers' attention since it was first introduced in the 1980s\cite{rumelhart1985learning}. In semi-supervised learning settings, they have a considerable volume of unlabelled data but only a small number of labeled training samples. In this setting, some study~\cite{rasmus2015semi} shows that  skip connections and layer-wise unsupervised targets effectively turn
autoencoders into hierarchical latent variable models, which are well suited for semi-supervised learning.
Because of its nature of clustering, it has been shown effective as an auxiliary task for many semi-supervised learning tasks such as regression \cite{guo2017deep}, and classification \cite{luo2017convolutional}. Based on the success of AE,  denoising autoencoders
(DAE)\cite{vincent2010stacked} and variational autoencoders (VAE)\cite{kingma2013auto} are  proposed for better representation learning. However, it has become less popular as more sophisticated methods have been proposed in recent years. 

\begin{figure}
\centering

\begin{tabular}{ c @{\hspace{10pt}} c }
    \includegraphics[width=.4\columnwidth]{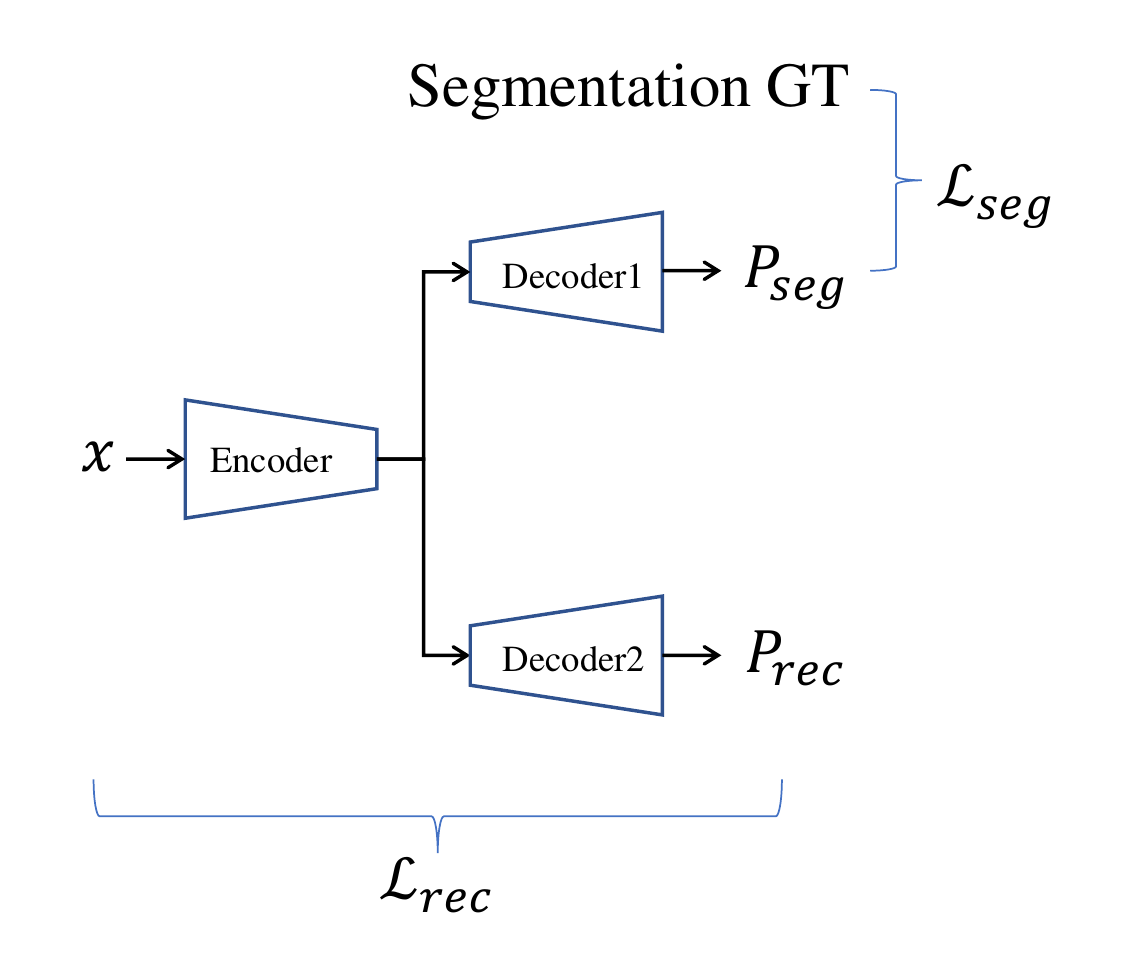}   &
      \includegraphics[width=.4\columnwidth]{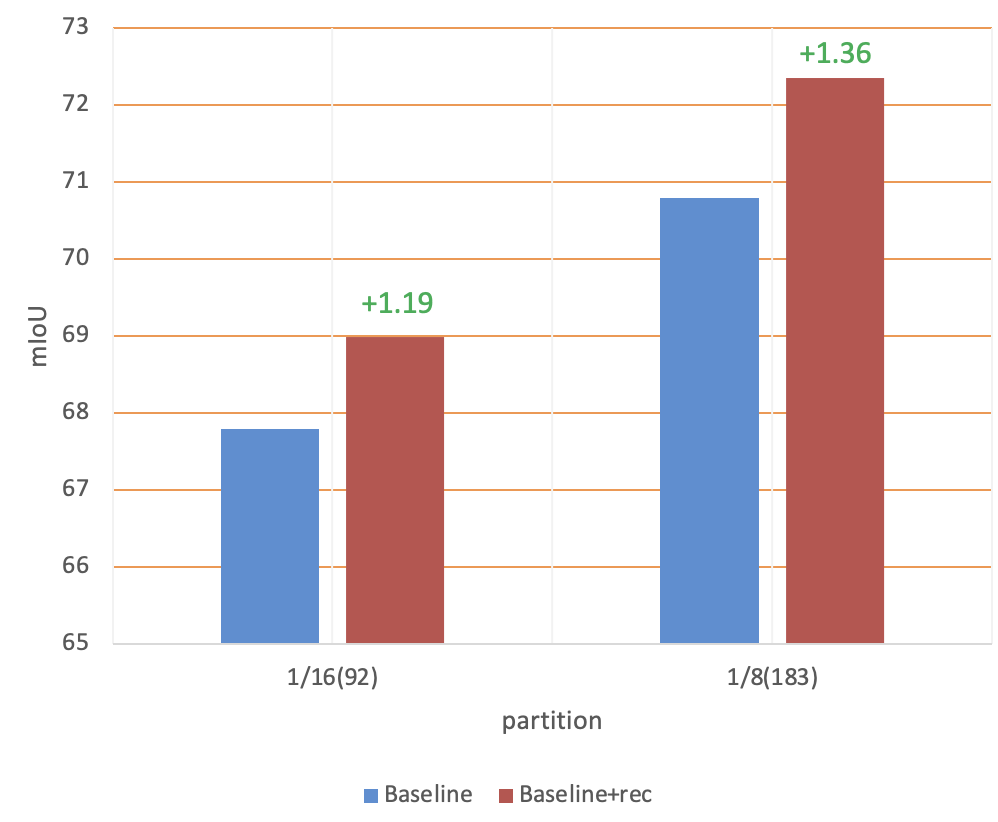}   \\
    \small (a) &
      \small (b)
  \end{tabular}

% \begin{subfigure}[0.23\textwidth]
%     \centering
%     \includegraphics[width=0.35\linewidth]{images/left-cropped.pdf}  
%     \caption{}
%     \label{toymodel}
% \end{subfigure}%
% \begin{subfigure}[0.23\textwidth]
%     \centering
%     \includegraphics[width=0.35\linewidth]{images/rec_his.png}  
%     \caption{}
%     \label{line0}
% \end{subfigure}
\caption{(a) illustrates a simple diagram of our joint training process. The model is based on the traditional teacher-student structure, and we add another decoder with the same architecture as the segmentation decoder to reconstruct the original image. When training, the optimization goal is to minimize the total loss from both segmentation CE loss and reconstruction MSE loss. In (b), we compare the performance of baseline with and without original image reconstruction on Pascal VOC 2012\cite{everingham2015pascal} under the standard partition protocols. It is worth noting that even by simply reconstructing the input image, the baseline outperforms current SOTA results in all partitions.}
\label{ab}
\end{figure}

\subsection{Supervised Semantic Segmentation}
As the fundamental task in computer vision, semantic segmentation has witnessed an explosion of progress in architecture design during recent decades. Starting from the FCN \cite{long2015fully}, which modifies the end-to-end architecture to be fully convolutional layers. After that,  some extensions were explored: 1) encoder-decoder structure\cite{badrinarayanan2017segnet,chen2018encoder,ronneberger2015u},  2) multi-scale aspects of the image \cite{chen2016attention,lin2016efficient},  3) pyramidal feature maps\cite{zhao2017pyramid}, 4) dilated convolutions \cite{chen2017deeplab,chen2018encoder,yu2015multi}. Recently, attention mechanisms\cite{fu2019dual,chen2016attention,huang2019ccnet} are popular among researchers because of their strength in global context communication. However, these fully supervised segmentation networks are data-hungry, laborious, and time-consuming.

\begin{figure*}[h!]
\begin{center}
  \includegraphics[width=1\linewidth]{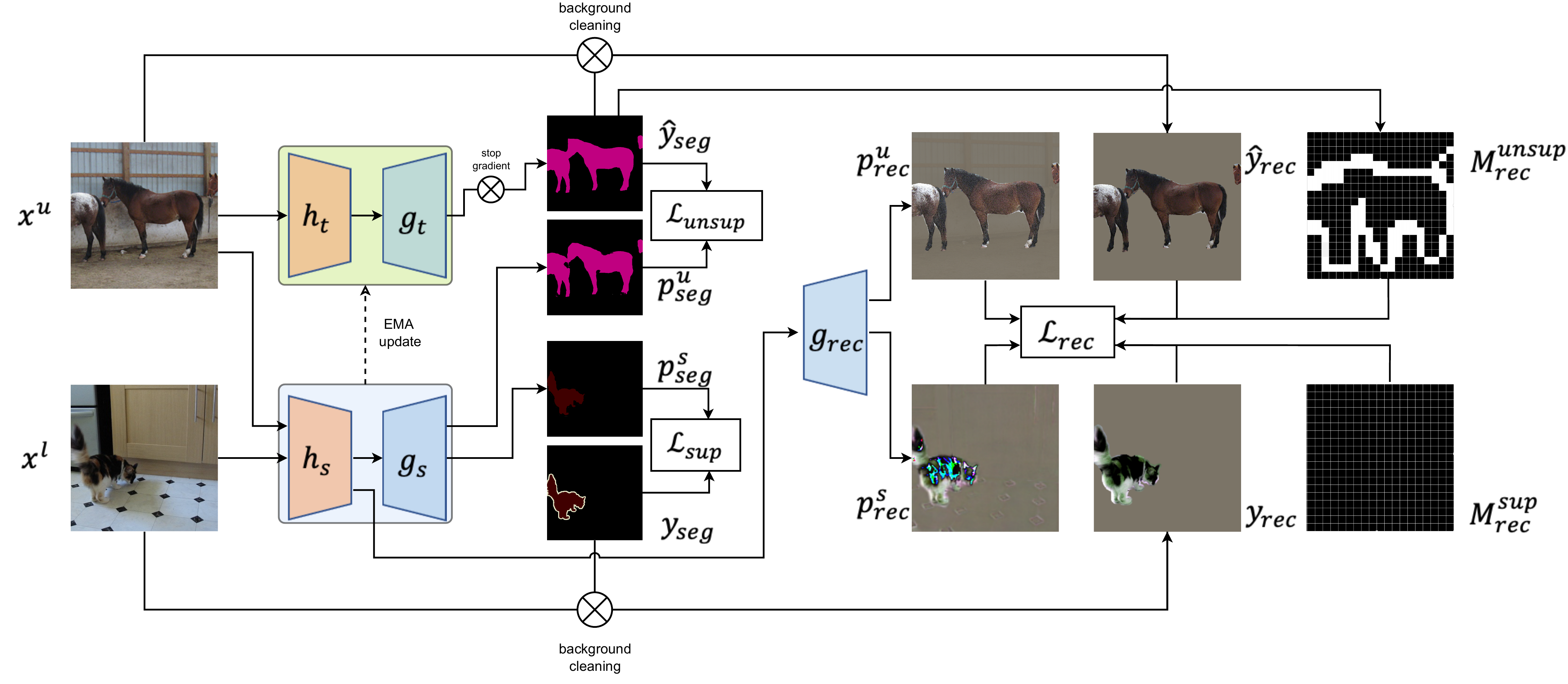}
\end{center}

  \caption{Illustration of our approach. The left side is the basic MT structure which employs two segmentation networks, the student $h_s,g_s$ and the teacher $h_t,g_t$. 
  The right side $g_{rec}$ is our Foreground-Only reconstruction module, which takes the output of the student's encoder $h_s$ and reconstructs a monotonous background image. Reconstruction loss masks are applied for loss calculation. Specifically, for labeled images, we set the background to 0 according to the ground truth, and the one-hot mask $M_{rec}^{sup} $ in shape [${H \times W \times 1}$] is all ones. For unlabeled images, the mask $M_{rec}^{unsup}$ is derived from the teacher’s ignoring uncertain areas. Then the teacher is updated by the student's exponential moving average (EMA).}
  \label{fig:model}
\end{figure*}
\subsection{Semi-supervised Semantic Segmentation}
Semi-supervised semantic segmentation aims to utilize the tremendous unlabeled and small amount of labeled data fully. Though the emergence of attention mechanisms for the model attracts much attention in the segmentation field,  SOTA semi-supervised semantic segmentation still relies on the CNN model Deeplab V3+~\cite{chen2017rethinking} and PSPNet~\cite{zhao2017pyramid}. One of the most existing studies used approaches is consistent learning with various perturbations\cite{yun2019cutmix,ouali2020semi,ke2020guided, liu2021perturbed}. For example, cutmix-seg \cite{yun2019cutmix} validates the effectiveness of image-level perturbation with cutmix data augmentation. Cross-consistency training (CCT)\cite{ouali2020semi} introduces a feature-level perturbation and constrains the outputs of different decoders. Similarly, guided collaborative training (GCT)\cite{ke2020guided} proposed a model-level perturbation with different network initialization and enforced the consistency between models. Most recently, PS-MT\cite{liu2021perturbed} introduced a new adversarial perturbation for double teachers for better prediction accuracy.

Another frequently used technique is self-training\cite{yuan2021simple,chen2021semi,yang2021st++}, which generates pseudo labels with unlabeled data and trains the model with labeled and pseudo-label data. Furthermore, considering the class imbalance and unreliable pseudo labels, recent studies \cite{hu2021semi,guan2022unbiased,wang2022semi} were proposed and achieved state-of-the-art performance.

\section{Image Reconstruction}
% formal formulation of reconstruction
Autoencoder~\cite{hinton2006reducing} is one of the oldest unsupervised/self-supervised learning approaches. It is usually implemented by an encoder-decoder pair. The encoder encodes the input image information into latent variables that are often low-dimensional or low-resolution. Then the decoder decodes the latent variables into an image of the same size as the input. A loss function is used to ensure the reconstructed output image is as close as possible to the input image, that is, 
\begin{align}
    \min_{\theta_e,\theta_d} \ \mathbb{E} \Bigl(\|x - f_{\theta_d}(f_{\theta_e}(x))\|^2_2\Bigl), 
\end{align}where $\mathbb{E}$ is a dissimilarity or distortion function and $f_{\theta_d}$, $f_{\theta_e}$ are the decoder and encoder, respectively.

Modern image segmentation neural networks, such as the DeepLab family \cite{papandreou15weak, florian2017rethinking, chen2017deeplab} can also be seen as an encoder-decoder structure, although the decoder is usually lightweight compared to the decoder. The encoder encodes the images into a $H' \times W' \times d$ dimensional feature map, and the decoder, e.g., ASPP module decodes the feature maps into the predicted segmentation mask. Therefore, the image reconstruction task can be naturally applied to the existing image segmentation neural networks by sharing the same encoder but using a different decoder (branch) to produce reconstructed images. 

\section{Image Reconstruction for Semi-supervised Semantic Segmentation}
%In this section, we will show that image reconstruction can be constructed as an effective auxiliary task for semi-supervised semantic segmentation. 
\subsection{Preliminary}
Semi-supervised semantic segmentation task is defined as:  given labeled images $X_l \in \mathbb{R}^{H \times W \times 3}$ , the corresponding pixel-wise semantic labels $y \in (1,C)^{H \times W} $, and unlabelled images $X_u \in \mathbb{R}^{H \times W \times 3}$ (W, H, C denotes the width, height, and the number of classes respectively). The goal is to learn a model $F$ from both label data $D^l = \{X_l,y\}$ and unlabelled data $D^u = \{X_u\}$. In most work \cite{yun2019cutmix,ouali2020semi,ke2020guided, liu2021perturbed,wang2022semi}, the overall optimization target is formalized as
\[
\mathcal{L} = \mathcal{L}_s +\lambda \mathcal{L}_{ul},
\]
where $\mathcal{L}_s$ and $\mathcal{L}_{ul}$ are loss functions for labeled and unlabeled images respectively.

\subsection{Semi-supervised Segmentation Baseline}
\label{sebsec:baseline}
Most of the state-of-the-art semi-supervised semantic segmentation methods \cite{yuan2021simple,chen2021semi,yang2021st++} are built around a simple baseline, as we call robust pseudo-labeling. Specifically, the model is firstly trained on the small number of labeled images and then produces the posterior probability estimation for each pixel on unlabeled images. Pseudo-labels are then generated if the highest posterior probability exceeds a predefined threshold. In robust pseudo-labeling, certain data augmentation, e.g., cutout \cite{devries2017improved}, cutmix \cite{yun2019cutmix}, is applied to the input image, and the pseudo-labels will be used to update the model with the augmented input images. 

Our semi-supervised segmentation baseline is based on a particular version of the robust pseudo-labeling approach. Specifically, our baseline follows the typical student-teacher framework in semi-supervised semantic segmentation \cite{ouali2020semi,ke2020guided, liu2021perturbed,wang2022semi}, with the teaching network parameters being the exponential moving average \cite{tarvainen2017mean} of the parameters of the student network.

Each network consists of a convolutional feature encoder $h$ and a segmentation decoder $g$. We denote the student version and teacher version of the encoder and decoder as $h_s$,$g_s$, $h_t$ $g_t$, respectively. At each training step, we equally sample $b$ labeled images $\mathcal{B}_l$ and $b$ unlabeled images $\mathcal{B}_{ul}$. For $\mathcal{B}_l$ and  $\mathcal{B}_{ul}$, we apply strong augmentations~\cite{chen2021semi, ke2020guided} (e.g., color jitter, randomize grayscale, blur, CutMix~\cite{yun2019cutmix} and zoom in/out~\cite{lin2018multi, chen2016attention}) to the student model. The teacher model will generate posterior distribution $P(y_{i,j}=c|I^n_{i,j})$, indicating the likelihood of each pixel $(i,j)$ being assigned to class $c$. A pseudo-label for pixel $(i,j)$ is generated if $\max_c P(y_{i,j} =c|I^n_{i,j}) > \tau$. Then the pseudo-labels will be used to train a student network with augmented input images.

\subsection{Reconstruction as an Auxiliary Task}
% say your findings fig1
% if mix clean+cutmix work, list formulations for l & ul respectively
Based on the framework mentioned in the Section \ref{sebsec:baseline}, we further incorporate image reconstruction as an auxiliary task, which is shown in Figure \ref{toymodel}. Different from the baseline, our framework has two decoders ($g_s$,$g_{rec}$) in the student network, and the two decoders share the encoder part ($h_s$).

Therefore, the outputs of the model are segmentation prediction: $P_{seg} = g_s \circ h_s(x) \in \mathbb{R}^{ H \times W  \times C}$ and the image reconstruction pixel value prediction $I_{rec} = g_{rec} \circ h_s(x) \in \mathbb{R}^{ H \times W \times 3 }$.
Same as \cite{chen2019multi}, we adopt the Mean Squared Error (MSE) loss for image reconstruction, and the image reconstruction module does not need additional annotations. The overall loss is defined as

\[
\mathcal{L} = \mathcal{L}_s +\lambda_1 \mathcal{L}_{ul} +\lambda_2 \mathcal{L}_{rec}.
\]
Surprisingly, this embarrassingly simple baseline achieves quite good performance. Figure \ref{ab} shows the performance before and after adding the reconstruction task. As seen, the benefit of using a reconstruction task is especially pronounced when the number of labeled images is small. 

\subsubsection{Visualization the ``Latent Images'' from the Reconstruction Decoder}
\label{sec:observation}
To understand the improvement, we perform visualization analysis on the reconstruction decoder. In particular, we consider the latent activations (feature maps) before the last layer of the reconstruction decoder. Recall that this last layer is a (kernel size $=1\times 1$) convolutional layer, which maps a $\mathbf{Z} \in \mathbf{R}^{H\times W \times d}$ feature map into the reconstructed image $I_{rec} \in \mathbb{R}^{H\times W \times 3}$. There are three convolutional filters $\mathbf{w}_1, \mathbf{w}_2, \mathbf{w}_3 \in \mathbb{R}^d$, one for each color channel. Now consider one output channel from $I_{rec}$, denoted as  $I^c_{rec}$,  it is clear that it can be written as
\begin{align}
I^c_{rec} = \sum_{k=1}^d \mathbf{Z}^k w^k_c,
\end{align}where $w^k_c$ denotes the $k$-th dimension of $\mathbf{w}_c$ and $\mathbf{Z}^k \in \mathbb{R}^{H \times W}$ denotes the $k$-th slice of $\mathbf{Z}$. Intuitively, the above equation suggests that the reconstructed image is the weighted average of $d$ slices $\mathbf{Z}^k$, where each $\mathbf{Z}^k$ is equivalent to an image, and we call it ``latent image''.

An interesting discovery is that those ``latent images'' could correspond to the semantic concepts in images if the encoder is jointly trained with a segmentation model. Figure \ref{fig:quality_results} shows some example ``latent images'' from an autoencoder trained with reconstruction loss only and trained jointly with semi-supervised segmentation loss. Also, to show the relative progress of the segmentation decoder and the reconstruction decoder, we choose an epoch that the training process has not converged yet. From Figure \ref{fig:quality_results}, we can make the following observations:
\begin{itemize}
    \item If the autoencoder is trained without the semi-supervised segmentation loss, the ``Latent Images'' have a weak correlation to the semantic concepts.
    \item When the autoencoder is trained with the semi-supervised segmentation loss, some ``Latent Images'' can correspond to some semantic concepts. For example, in the second and fourth row of Figure \ref{fig:quality_results}, the activation areas in some feature maps are focused on the train and the sheep. It seems that the semantic segmentation loss provides an inductive bias to make the image reconstructed through semantically meaningful ``latent images''. 
    \item Surprisingly, some ``Latent Images'' recover the object contour better than the segmentation decoder at the same training epoch. It seems that the ``Latent Images'' are leading the segmentation decoder, which might explain why the reconstruction task could help segmentation.
    \item Finally, we find the ``Latent Images'' are far from perfect. Some background pixels, especially those that are the context of the object, tend to co-occur with the object in the latent image.
\end{itemize}

\subsection{Improving the Reconstruction Task by Object-Background Disentanglement}
% background --> 0
% According to our findings above, we further proposed a monotonous background reconstruction (MoBre) to replace the standard image reconstruction task to disentangle the
% object clue from the background patterns

% For labeled image $\mathcal{B}_l$, instead of reconstructing the original input image, here we firstly obtain the reconstruction target (foreground pixels) by converting to the monotonous background for each input image $x_l \in \mathcal{B}_l$:
The last observation discussed in Section \ref{sec:observation} suggests a potential object-background entanglement may exist in the current reconstruction task. Thus, we propose the following strategy to disentangle the object and its context background. More specifically, we let the reconstruction decoder only reconstruct the foreground-only images at the labeled set. In other words, the output from the reconstruction decoder only contains pixels belonging to the object parts while the background pixels are set to zero, that is,
\begin{equation}  \label{equation3}
y_{rec}\  _{(i,j)}= 
\begin{cases}
    x^l_{(i,j)},& \text{if}\  y_{(i,j)}\in \text{foreground}\\
    0,              & \text{otherwise}
\end{cases}
\end{equation}

Examples of foreground-only images are shown in Figure \ref{fig:model}. For unlabeled images, we do not have access to the class labels, those we cannot directly generate foreground-only images. Thus we recourse to pseudo-labels. For an unlabeled image, we consider the following three scenarios for a pixel $(i,j)$: (1) the current pixel can generate a pseudo-label, and the pseudo-label corresponds to the foreground. In other words, $max_{c'\in \mathcal{O}} P(y_{i,j} = c'|x_{i,j}) > \tau$, where $\mathcal{O}$ is the set of classes that belong to objects. (2) the current pixel can generate a pseudo-label, and the pseudo-label corresponds to the background. (3) no pseudo-label can be generated from the current pixel and the segmentation decoder is uncertain about the class of the pixel. We will ignore the loss of pixels from (3) and only perform the foreground-only reconstruction for pixels from (1) and (2).

We call this modified reconstruction method as Foreground-Only reconstruction (FOrec). The scheme is illustrated in Figure \ref{fig:model}.

\noindent \textbf{Discussion:} The FOrec method is mainly for object centric semantic segmentation, where the aim is to segment different objects from the background. For generic scene segmentation, i.e., segmenting both things and stuff, one can choose a category that often co-occurs with other categories as the background. Applying FOrec in that case could potentially alleviate the entanglement of those semantic concepts. 

\section{Experiments}
In this section, we compare our approach with several semi-supervised semantic segmentation methods.
%----------------------------------------------------------------------------------------------------------------------------------------
\subsection{Experimental Setup}
\textbf{Datasets:}  
Our experiments are mainly conducted on the Pascal VOC 2012 \cite{everingham2015pascal}, and Cityscapes \cite{cordts2016cityscapes}, which are widely used in semi-supervised semantic segmentation tasks \cite{yun2019cutmix,ouali2020semi,ke2020guided, liu2021perturbed}. The \textbf{classic} Pascal VOC 2012 consists of 1,464/1,449/1,556 images covering twenty classes for training, validation, and testing, respectively. Due to the demand for data for the semi-supervised semantic segmentation scene, some researchers \cite{ouali2020semi,ke2020guided, liu2021perturbed} adapt the additional labels from \cite{hariharan2011semantic}, which means the training data is augmented up to 10,528 images. In the augmented training set, 1,464 labeled data is selected among 1,464 samples in the \textbf{classic} setting, while the remainings are of low quality, containing noise. The augmented data selection setting is named \textbf{blender}. Both settings are evaluated for the measurement of performance.
Cityscapes \cite{cordts2016cityscapes}  is the urban driving scene dataset, consisting of 2,975, 500, and 1,525 images covering 19 classes for training, validation, and testing,  respectively.

In this paper, we follow the same data splitting protocol from U$^2$PL\cite{wang2022semi} and experiment with four kinds of label partition: 1/16, 1/8, 1/4, and 1/2. Our code will be released after the anonymity period.

\textbf{Evaluation metrics:} Following the previous works \cite{wang2022semi,liu2021perturbed}, we adopt the mean Intersection-over-Union (mIoU) as the evaluation metrics.

\textbf{Implement detail:} Following the prior work~\cite{liu2021perturbed,xu2022semi,chen2021semi}, the network structure of our method is based on Deeplab V3+~\cite{chen2017rethinking}, with pretrained ResNet-101 as the backbone. The segmentation head and the auxiliary task head are the default pixel-level linear classifier.

For all experiments on both datasets, we employ the stochastic gradient descent (SGD) as the optimizer and polynomial learning rate decay: $(1-\frac{iter}{total_iter})^{0.8}$ for model optimization. For reconstruction loss, we set the $\lambda_1$ as 0.5  and $\lambda_2$ as 1 for the unsupervised and supervised parts, respectively.

On Pascal VOC 2012 \cite{everingham2015pascal}, the images are cropped into $512 \times 512$ pixels and trained with initial learning rate  $1.0 \times 10^{-3}$ 
, weight decay $1.0 \times 10^{-4}$ ,and 80 training epochs.   On Cityscapes \cite{cordts2016cityscapes} , we crop the images into $712 \times 712$ pixels and trained our model with an initial learning rate $1.0 \times 10^{-2}$ 
, weight decay $5.0 \times 10^{-4}$ and 200 training epochs. 
Our experiments were run with batch size 16 on 8 NVIDIA Tesla V100 GPUs.  

For the reconstruction decoder, we simply adapt the same structure as the segmentation decoder. The only difference between them is the channel number of the last layer.

\subsection{Comparison with State-of-the-Arts}
\subsubsection{Pascal VOC 2012}
\subsubsection{Pascal VOC 2012}
Table \ref{table:classic} and Table  \ref{table:blender} illustrated the results on Pascal VOC 2012 validation set, Table  \ref{table:classic} is under \textbf{classic} setting and Table \ref{table:blender}  is under \textbf{blender} setting.

For \textbf{classic} setting, Table  \ref{table:classic} illustrates that our approach successfully exploits unlabelled data, with a dramatic performance boost from the fully supervised training. Specifically, in the smaller partition like  92 and 183, our approach surpasses the fully supervised baseline with 25.2\% and  19.8\%, respectively. Meanwhile, compared with the current SOTA methods, our approach performs consistently better than all other methods for all partition protocols (using the ResNet-101 as the backbone). Taking the U$^2$PL\cite{wang2022semi} as the instance, our approach improves the performance by 1.6\% to 5.6\% in all cases. Furthermore, in some partitions, our approach is better than all the current SOTA methods with fewer labeled samples. For example, compared with 
all current SOTA methods, our approach trained with 92 labeled images outperforms trained with 183 labeled images. This demonstrates that when the number of labeled data is extremely small (92, 183), our approach achieves a significant improvement in performance. 

\begin{table}[h]

\centering
\begin{tabular}{l|ccccc}
\toprule
 Method  &92  &183  &366  &732  &1464  \\ 
 \midrule
 Supervised &45.8  &54.9  &65.9  &71.7  &72.5  \\ 
 \midrule
 MT \cite{tarvainen2017mean} &51.7  &58.9  &63.9  &69.5  &71.0  \\
 PseudoSeg \cite{zou2020pseudoseg}  &57.6  &65.5  &69.1  &72.4  &73.2  \\
 CPS \cite{chen2021semi}{\tiny [CVPR 21'] }  &64.1  &67.4  &71.7  &75.9  &-  \\
 PS-MT\cite{liu2021perturbed}{\tiny [CVPR 22']} &65.8  &69.6  &76.6  &78.4  &80.0  \\ 
 ST++\cite{yang2021st++}{\tiny [CVPR 22']} &65.2 &71.0  &74.6  &77.3  &79.1  \\ 
 U$^2$PL\cite{wang2022semi}{\tiny [CVPR 22'] } &68.0  &69.1  &73.6  &76.2  &79.5  \\
 % SS-proto\cite{xu2022semi}{\tiny [NIPS 22']} &70.1  &\textbf{74.7}  &77.2  &78.5  &80.6  \\ 
 \midrule
 FOrec (Ours) &\textbf{71.0}  &\textbf{74.7}  &\textbf{77.5}  &\textbf{78.7}  &\textbf{81.1}  \\ 
 \bottomrule
\end{tabular}

\caption{Comparing results of state-of-the-art algorithms on PASCAL VOC 2012 \cite{everingham2015pascal} val set with mIoU
(\%)  metric. Methods are trained on the classic setting, i.e., the labeled images are selected from the
original VOC train set, which consists of 1, 464 samples in total. Best results are in bold.}
\label{table:classic}
\vspace{-10pt}
\end{table}

For \textbf{blender} setting, Table  \ref{table:blender} indicates that our approach outperforms the supervised baseline by a large gap 3.86\% to 10.97\% from the fully supervised training. 
Compared with the current SOTA methods, our approach beats all other methods for all partition protocols. Even comparing with one of the currently best performed approaches U$^2$PL\cite{wang2022semi}, our approach improves the performance by 1.63\%, 1.51\%, 1.62\% in 662, 1323, and 2646, respectively.  The impressive boost proves that our method is not only useful for accurately annotated data, but also compatible with the noisy annotation.

\begin{table}[h]
\centering
\begin{tabular}{l|cccc}
\toprule
 Method  &662  &1323  &2646  &5290   \\ 
 \midrule
 Supervised &67.87 &71.55 &75.80  &77.13\\
 \midrule
 MT\cite{tarvainen2017mean} &70.51  &71.53 &73.02 &76.58  \\
 CPS \cite{chen2021semi}{\tiny [CVPR 21']} &74.48  &76.44  &77.68  &78.64   \\
 AEL\cite{hu2021semi} {\tiny [NIPS 21']} &77.20 & 77.57 &78.06  &80.29  \\
 ST++ \cite{yang2021st++}{\tiny [CVPR 22']} &74.70 &77.90  &77.90 &-   \\
 PS-MT \cite{liu2021perturbed}{\tiny [CVPR 22']} &75.50  &78.20  &78.72  &79.76   \\ 
 UCC \cite{fan2022ucc}{\tiny [CVPR 22']} &76.49  &77.06  &79.09  &79.54   \\
 U$^2$PL\cite{wang2022semi} {\tiny [CVPR 22']} &77.21 &79.01 &79.30 &80.50   \\
 
 % SS-proto \cite{xu2022semi}{\tiny [NIPS 22']} &78.60  &\textbf{80.71}  &80.78  &80.91   \\
 \midrule
 FOrec (Ours) &\textbf{78.84}  &\textbf{80.52}  &\textbf{80.92}  &\textbf{80.99}    \\ 
 \bottomrule
\end{tabular}
\caption{Comparing results of state-of-the-art algorithms on PASCAL VOC 2012 \cite{everingham2015pascal} val set with mIoU
(\%) metric. Methods are trained on the blender setting, i.e., the labeled images are selected from
the augmented VOC train set, which consists of 10, 582 samples in total.  Best results are in bold.}
\label{table:blender}
\vspace{-10pt}
\end{table}

\subsubsection{Cityscapes}
Table~\ref{table:cityscape} demonstrates the results of our method against several current state-of-the-art algorithms  on Cityscapes validation set. Compared to the fully supervised results, our
method successfully exploits unlabelled data, with an obvious performance boost for all partitions. e.g., under the 1/16 label partition, our approach surpasses the fully supervised result by 6.68\%. Then, compared
to the state-of-the-art algorithm U$^2$PL\cite{wang2022semi}, Ours performs better than U$^2$PL in all cases  by 2.12\%, 1.39\% and 1.18\% under the 1/16, 1/8 and 1/4 label partition, respectively.

Note that the performance of our method on the 1/16 label partition is slightly lower than that of AEL~\cite{hu2021semi}. The reason is that the class imbalance problem of this partition is more serious and AEL especially aims to deal with class imbalance problems. However, our methods focus on separating the foreground objects from the background
patterns in semi-supervised semantic segmentation tasks, we do not explicitly consider processing label imbalance problems. Technically, there is a high probability that merging both ideas is a to optimize overall performance.
\begin{table}[h]
\centering
\begin{tabular}{l|cccc}
\toprule
 Method  &1/16  & 1/8 & 1/4  & 1/2 \\ 
 \midrule
 Supervised &65.74 &72.53 &74.43 & 77.83 \\ 
 \midrule
 MT\cite{tarvainen2017mean} &69.03  &72.06 &74.20 &78.15 \\
 CCT \cite{ouali2020semi}{\tiny [CVPR 20']} &69.32  &74.12 &75.99  &78.10  \\
 CPS \cite{chen2021semi}{\tiny [CVPR 21']} &69.78  &74.31  &74.58  &76.81  \\
 AEL \cite{hu2021semi}{\tiny [NIPS 21']} &\textbf{74.45}  &75.55  &77.48  &79.01   \\ 
 U$^2$PL\cite{wang2022semi} {\tiny [CVPR 22']} &70.30 & 74.37 &76.47 &79.05   \\\hline
 FOrec (Ours) &72.42  &\textbf{75.76}  &\textbf{77.65}  &\textbf{79.18}    \\
 \bottomrule
\end{tabular}
\caption{Comparing results of state-of-the-art algorithms on Cityscapes \cite{cordts2016cityscapes} val set with mIoU (\%) $\uparrow$
metric. Methods are trained on identical label partitions, and the labeled images are selected from the Cityscapes train set, which consists of 2, 975 samples in total. Best results are in bold.}
\label{table:cityscape}
\end{table}

\begin{table}[t]%\footnotesize
    \centering
    \begin{tabular}{l | c | c | c c }
         \toprule
         & \makecell[c]{Standard \\Reconstruction} & \makecell[c]{FOrec} & 1/16 (92) & 1/8 (183) \\
         \midrule
         \ding{172} & \xmark & \xmark & 67.82  & 70.78 \\
         \ding{173} & \cmark & \xmark & 68.99 & 72.35\\
         \ding{174} & \xmark & \cmark & \textbf{70.99} & \textbf{74.67} \\
         \bottomrule
    \end{tabular}
    \caption{Ablation study on the effectiveness of different components of our approach. 
    \cmark and \xmark \ represent the variant containing  or not containing the sub module at each row respectively.
    }
    \label{tab:abl_component}
    \vspace{0.1cm}
    % \vspace{-0.5cm}
\end{table}

% \begin{table}[t]%\footnotesize
%     \centering
%     \caption{Ablation study on the background selection in our approach for Cityscapes dataset. 
%     }
%     \label{tab:abl_bg_selection}
%     \begin{tabular}{  c | c c c c }
%          \toprule
%          & 1/16 (186) & 1/16 (372) & 1/4 (744) & 1/2 (1488)\\
%          \midrule
%           void &\textbf{72.42}  &\textbf{75.76}  &77.65  &\textbf{79.18}\\
%           road & 71.38 &75.38 & \textbf{77.85} &78.92\\
%          \bottomrule
%     \end{tabular}
% \end{table}

\begin{table}[h]\small
\vspace{-0.4cm}
\centering
\begin{tabular}{l|ccccc}
\toprule
 Method  &92  &183  &366  &732 \\ 
 \midrule
 %Baseline &67.8  &71.2  &75.6  &77.4 \\ 
 fore/back-ground seg. & 67.7 & 71.9   & 76.1  & 78.1  \\ 
 %Baseline+rec &69.0 &72.4   &76.3  &78.1  \\ 
 \textbf{FOrec (ours)} &\textbf{71.0}  &\textbf{74.7}  &\textbf{77.5}  &\textbf{78.7}    \\ 
 \bottomrule
\end{tabular}
\vspace{-0.3cm}
\caption{Foreground-background (saliency estimation) segmentation on PASCAL VOC 2012 (classic setting).}
\label{tab:abl_fbg_seg}
\vspace{-0.4cm}
\end{table}

\subsection{Analysis}
In the following part, we perform a series of experiments to analyze the proposed method. Specially, we consider the following analyses: (1) The comparison between FOrec and standard reconstruction on both the object-centric segmentation task, i.e., PASCAL VOC, and scene-understanding segmentation task, i.e., CityScapes. (2) The latent images created from FOrec. (3) The applicability of the proposed method to other semi-supervised segmentation approaches. 

\noindent \textbf{The comparison of the standard reconstruction and FOrec} 
In Table~\ref{tab:abl_component}, we compare the standard reconstruction task and the proposed foreground-only reconstruction scheme. As seen, FOrec achieves a significant improvement over the standard reconstruction task on PASCAL VOC 2012 clean setting, the improvement over the standard reconstruction is around 2\%. This supports our claim that using foreground-only images as the reconstruction target could be beneficial for object-centric segmentation. 

\noindent \textbf{The comparison of the foreground-background segmentation and FOrec} 
Table~\ref{tab:abl_fbg_seg} shows that the naive foreground-background segmentation performs worse than our approach,  the improvement benefit from the  foreground-only reconstruction is ranging from  0.5\% to 3.3\%.  We think that unifying all kinds of foregrounds into one class is not conducive to the semantic segmentation of different objects.
%There are mainly two components in our proposed approach, i.e., the reconstruction loss and the background cleaning module. Table~\ref{tab:abl_component} shows the performance contribution of each component on two kinds of labeled splits on PASCAL VOC 2012 clean setting. The first line presents the results of the pseudo-labeling baseline, and the second line shows that adding the default reconstruction loss can greatly improve the baseline. After incorporating the proposed background cleaning module, the final performance boosts further and achieves an overall best performance compared to existing semi-supervised semantic segmentation methods.

\begin{figure*}
\centering
\includegraphics[width=\linewidth]{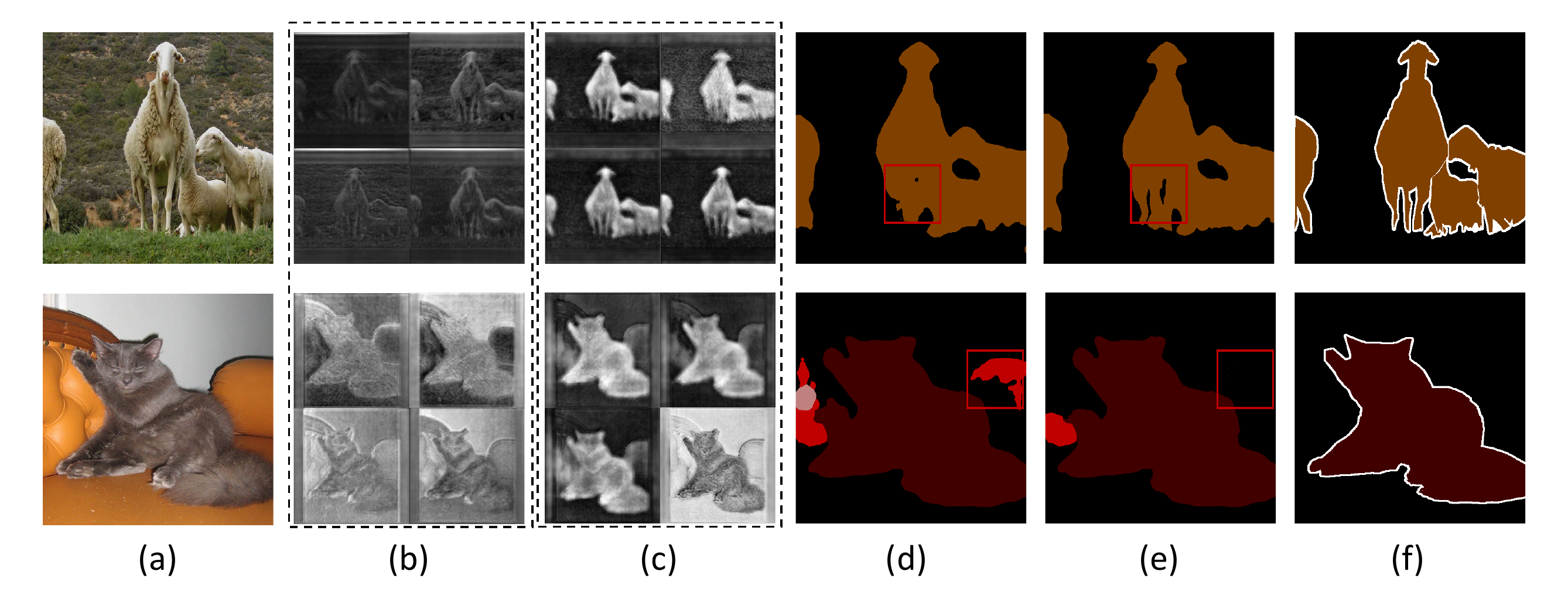}  

\caption{Visualizations of the latent images obtained from FOrec and standard reconstruction, from left to right, are (a) input images, (b) latent images obtained from standard reconstruction, (c) latent images obtained from FOrec, (d) segmentation results from standard reconstruction, (e) segmentation results from FOrec, and (f) segmentation ground truth. Note that the (b) and (c) are slices from the same position for a fair comparison.}

\label{final}
\vspace{-10pt}
\end{figure*}
% \begin{figure*}
%     \centering
%     \begin{subfigure}[b]{0.25\linewidth}
%         \includegraphics[width=\linewidth]{cvpr2023-author_kit-v1_1-1/latex/resize1.jpg}
%         \includegraphics[width=\linewidth]{cvpr2023-author_kit-v1_1-1/latex/resize2.jpg}
%         \caption{}
%      \end{subfigure}
%      \begin{subfigure}[b]{0.25\linewidth}
%         \includegraphics[width=\linewidth]{cvpr2023-author_kit-v1_1-1/latex/recon_0_255.png}
%         \includegraphics[width=\linewidth]{cvpr2023-author_kit-v1_1-1/latex/recon_bg0_255.png}
%         \caption{}
%      \end{subfigure}
%     \begin{subfigure}[b]{0.25\linewidth}
        
%         \includegraphics[width=\linewidth]{cvpr2023-author_kit-v1_1-1/latex/recon_16_255.png}
%         \includegraphics[width=\linewidth]{cvpr2023-author_kit-v1_1-1/latex/recon_bg0_16_255.png}
%         % \includegraphics[width=\linewidth]{latex/images/qualitive_results/206/t_u_0.pdf}
%         % \includegraphics[width=\linewidth]{cvpr2023-author_kit-v1_1-1/latex/hf_img31.png}
%     \caption{}
%     \end{subfigure}
    
%     \caption{Visualisation results from \textit{Pascal VOC 2012 \textbf{validation set}}. (a) input images, (b) feature maps from the pure reconstruction model, (c) feature maps from the reconstruction-segmentation model, and (d) ground truth. 
%     } \vspace{-10pt}
%     \label{fig:quality_results}
% \end{figure*}

\noindent \textbf{The impact of FOrec on the ``latent images''} The FOrec is proposed to address the issue that background pixels tend to co-occur with foreground pixels. To verify this design, we visualize the latent images obtained from FOrec and standard reconstruction. We conduct an experiment by using two models, one trained by FOrec and another trained by standard reconstruction (jointly trained with semi-supervised segmentation loss for both cases with the same architecture). Then we use the trained model to generate latent images for input images from the validation set. \textit{Note that although FOrec is trained to reconstruct foreground-only images, the Foreground-only images are only used at the training time as the target. Once trained, it can be used without knowing the foreground mask.} The results are shown in Figure~\ref{final}. As seen, by applying FOrec, the latent images tend to capture more object regions. We can observe more ``latent images'' that only include the objects. For example, in our figures, the goats and the cat in (c) are highlighted a lot compared with the latent images of standard reconstruction. When mapping to the results
of the segmentation task, these natures are kept. For instance,
in the first example in Figure~\ref{final}, the gaps between goats' legs are captured,  which did not appear in the results from
the standard reconstruction. Moreover, in the second example,
the sofa next to the cat is faded in the FOrec feature maps.
In the segmentation results, it disappears.  These results in Figure~\ref{final} clearly validate the effectiveness of the proposed method.

\begin{table}[h]
\centering

\begin{tabular}{l|ccccc}
\toprule
 Method  &92  &183  &366  &732 \\ 
 \midrule
 PS-MT\cite{liu2021perturbed}&65.8  &69.6  &76.6  &78.4 \\ 
 PS-MT+rec &68.4 &71.0   &77.2  &78.9  \\ 
 PS-MT+FOrec &\textbf{70.3} &\textbf{71.9}   &\textbf{77.9}  &\textbf{79.8}    \\ 
 \bottomrule
\end{tabular}
\caption{Comparing results of another SOTA codebase (PS-MT\cite{liu2021perturbed}) on classic setting PASCAL VOC 2012 \cite{everingham2015pascal} val set with mIoU
(\%)  metric. Experiments are conducted in the same settings as in the paper. }
\label{table:psmt}
\vspace{-10pt}
\end{table}
%----------------------------------------------------------------------------------------------------------------------------------------

\noindent \textbf{The applicability of the proposed method on other semi-supervised segmentation methods}
Finally, we apply both the reconstruction task and FOrec to another semi-supervised learning approach, PS-MT \cite{liu2021perturbed}. The results are shown in Table \ref{table:psmt}. As seen, the reconstruction is still effective. Compared with the PS-MT baseline, applying FOrec leads to a significant increase, especially when the number of training examples is small. The advantage of FOrec over standard reconstruction is also seen. Again, we observe FOrec tends to produce a superior performance on the low-supervision regime, e.g., when only 92 labeled images are used. 

% \section{Limitation}
% As the propose of our method is to disentangle the object clue from the background patterns, 
\section{Conclusion}

In this paper, we revisit the idea of using image reconstruction as an auxiliary task for semi-supervised semantic segmentation. We find that this old idea can produce results competitive with state-of-the-art semantic segmentation algorithms. By visualizing the intermediate layer activations of the image reconstruction module, we show that the feature map channel can correlate well with the semantic concept, which explains why joint training with the reconstruction task is helpful for the segmentation task. Motivated by this observation, we further proposed a modification to the image reconstruction task, aiming to further disentangle the object clue from the background patterns. From experiment evaluation on various datasets, we show that using reconstruction as an auxiliary loss leads to consistent improvements in various datasets and methods. The proposed method can further lead to significant improvement in object-centric segmentation tasks. For datasets without background class, it only provides slight improvements and more investigations will be made to improve scene understanding ability.
% One limitation of the current method is that it mainly works for object-centric segmentation. It only provides slight improvements for datasets without background class. More investigations will be made to improve reconstruction-based task for scene understanding oriented image segmentation.

{\small
\bibliographystyle{ieee_fullname}
\bibliography{egbib}
}

\end{document}